\documentclass[letterpaper, 10 pt, conference]{ieeeconf}  

\IEEEoverridecommandlockouts                              

\usepackage{amsmath}
\usepackage{algorithm}
\usepackage{algpseudocode}
\usepackage{cite}
\usepackage{comment}
\usepackage{amssymb}
\usepackage{booktabs}  
\usepackage{adjustbox} 
\usepackage{stfloats}
\usepackage{array}

\title{\LARGE \bf A Graph-Based Reinforcement Learning Approach with Frontier Potential Based Reward for Safe Cluttered Environment Exploration}

\author{Gabriele Calzolari, Vidya Sumathy, Christoforos Kanellakis, George Nikolakopoulos
\thanks{This work was partially supported by the Wallenberg AI, Autonomous Systems and Software Program (WASP) funded by the Knut and Alice Wallenberg Foundation, and by the European Union's Horizon Europe Research and Innovation Program, under the Grant Agreement No. 101119774 SPEAR.}
\thanks{This research was conducted using the resources of High Performance Computing Center North (HPC2N). Additionaly, the RL-training were enabled by resources provided by the National Academic Infrastructure for Supercomputing in Sweden (NAISS), partially funded by the Swedish Research Council through grant agreement no. 2022-06725.}
\thanks{The authors are within the Robotics and AI Group, Department of Computer Science, Electrical and Space Engineering, Luleå University of Technology, Sweden. Corresponding author's e-mail: gabcal@ltu.se}}

\begin{document}

\maketitle
\thispagestyle{empty}
\pagestyle{empty}

\begin{abstract}

Autonomous exploration of cluttered environments requires efficient exploration strategies that guarantee safety against potential collisions with unknown random obstacles. This paper presents a novel approach combining a graph neural network-based exploration greedy policy with a safety filter to ensure safe navigation goal selection. The network is trained using reinforcement learning and the proximal policy optimization algorithm to maximize exploration efficiency while reducing the safety filter interventions. However, if the policy selects an unfeasible action, the safety filter intervenes to choose the best feasible alternative, ensuring system consistency. Moreover, this paper proposes a reward function that includes a potential field based on the agent's proximity to unexplored regions and the expected information gain from reaching them. Overall, the approach investigated in this paper merges the benefits of the adaptability of reinforcement learning-driven exploration policies and the guarantee ensured by explicit safety mechanisms. Extensive evaluations in simulated environments, along with comparisons against established benchmarks, demonstrate that the proposed approach achieves efficient and safe exploration in cluttered settings.

\end{abstract}

\section{Introduction}

Leveraging robotic systems to achieve autonomous exploration of cluttered environments is of relevant importance for applications such as search and rescue, environmental monitoring, and infrastructure inspection. However, this task is challenging  due to the partial observability of the environment, sensor noise, and the presence of obstacles that can make reliable mapping and localization more difficult. Furthermore, narrow traversable regions and dynamic obstacles complicate motion planning and collision avoidance under uncertainty so that exploration frameworks require a balance between exploration efficiency and safety under real-time constraints. Recently, Reinforcement Learning (RL) has been increasingly used to design policies to enable autonomous agents to explore these environments \cite{garaffa2021reinforcement, AZPURUA2023104304, QuattriniLi2020, wang2025multirobotcooperativeexplorationunknown}. \cite{xue2022uav} proposes an RL-based navigation framework for UAVs employing the fast recurrent stochastic valued gradient algorithm for safe exploration in cluttered 3D environments. \cite{cimurs2021goal} integrates TD3-based DRL with global waypoint selection for autonomous goal-driven exploration without prior maps or human input. \cite{cao2023ariadne} proposes ARiADNE, an attention-based RL approach that enables real-time, non-myopic path planning by learning spatial dependencies and predicting exploration gains. \cite{sun2022fully} proposes ReLMM that autonomously teaches robots navigation and grasping via RL, using modular policies, uncertainty-driven exploration, and autonomous resets. \cite{11128566} introduces a novel hierarchical integration of a GNN-based exploration policy with an explicit safety filter that guarantees collision-free actions, enabling reliable deployment in real-world cluttered environments while maintaining exploration efficiency. \cite{calzolari2025platformagnosticreinforcementlearningframework} investigates a decentralized RL framework that jointly optimizes exploration and communication through a constrained, proximity-based information-sharing strategy and a density-aware frontier representation, improving scalability and reducing redundant exploration in multi-agent systems. 
In particular, RL policies are typically parameterized by deep neural networks and therefore lack interpretability and provide no formal guarantees of correctness beyond empirical evaluation. This poses challenges in safety-critical applications requiring satisfaction of strict constraints. Recent work addresses this by integrating explicit safety mechanisms enforcing hard action constraints into RL frameworks, thereby improving robustness and enabling more reliable real-world deployment \cite{gu2024review, brunke2022safe, zhao2023state, xu2022trustworthy, guerrier2024learning, ma2022conservative, yang2022cup, zhang2023evaluating}. \cite{dai2023augmented} introduces a novel primal–dual safe RL framework that stabilizes training and achieves precise constraint satisfaction by augmenting the Lagrangian with a quadratic penalty, enabling efficient optimization with first-order methods and improved convergence over prior approaches. \cite{s25175488} introduces a novel risk-aware safe RL framework that integrates PPO with control barrier functions to enforce real-time safety via optimization-based filtering, enabling adaptive and efficient collision avoidance in complex dynamic environments. \cite{10496239} introduces a hybrid safe multi-agent motion planning framework that combines RL-based trajectory generation with convex optimization and chance-constrained control to enforce probabilistic safety guarantees, enabling real-time collision-free coordination in uncertain and cluttered environments.

\begin{figure*}[h]
    \centering
    \includegraphics[width=\linewidth]{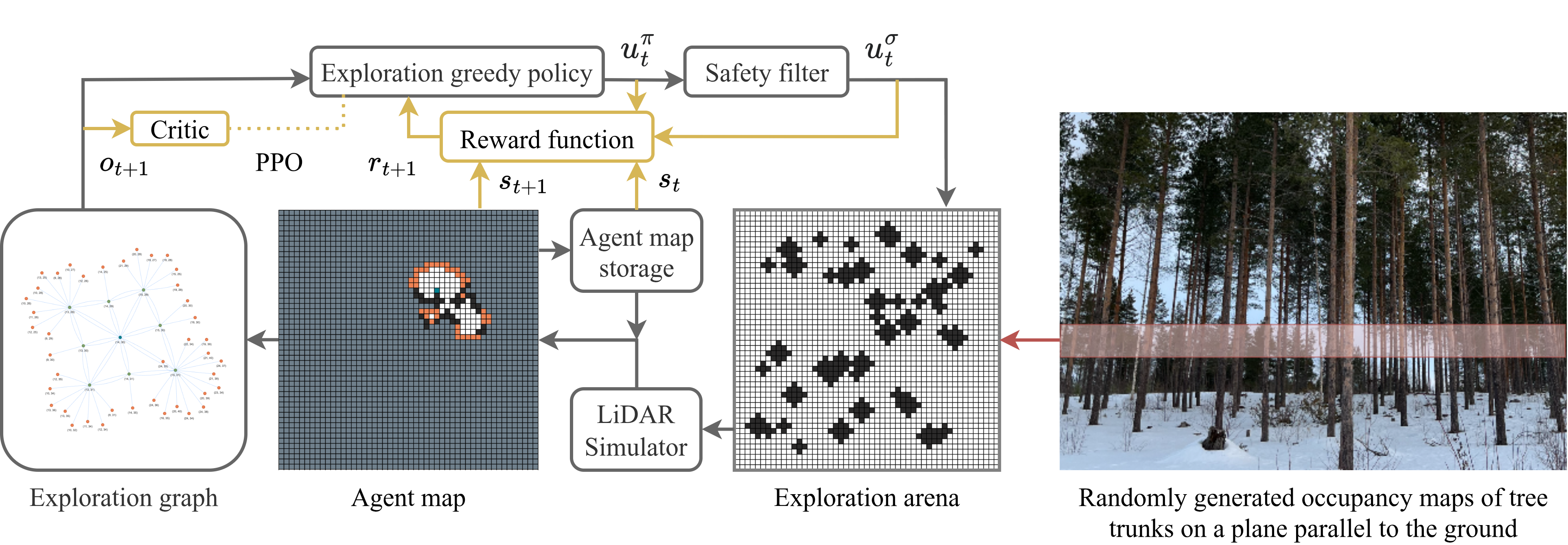} 
    \caption{Overview of the proposed safe reinforcement learning framework for exploring unknown, cluttered arenas, using randomly generated occupancy maps to model tree trunk distribution on a plane parallel to the ground. The image highlights key elements required during training (yellow and gray) and execution (gray).}
    \label{fig:framework_overview}
\end{figure*}

Among policy architectures, Graph Neural Networks (GNNs) have gained increasing attention due to their ability to encode spatial relationships through graph representations, enabling policies to extract and exploit structurally relevant features for informed navigation decisions \cite{liu2023graph, munikoti2023challenges, nie2023reinforcement}. \cite{zhang2021autonomous} introduces a novel graph-based spatiotemporal DRL framework (Graph-STNN) that integrates GCN, TCN, and GRU to jointly capture spatial structure, temporal dynamics, and historical information in exploration tasks, enabling more effective and generalizable target selection in unknown environments compared to existing methods. \cite{herrera2023learning} proposes a novel GNN-based reinforcement learning framework for autonomous exploration that learns an efficient exploration policy directly on graph representations, enabling scalable decision-making and strong generalization across varying graph topologies while achieving performance comparable to state-of-the-art planning methods.

Considering the literature, we propose a safe autonomous exploration method for cluttered forest-like environments, where an agent moves on a 2D plane at fixed altitude between tree trunks. The main contributions of this paper are threefold. First, we introduce a reinforcement learning framework that combines a graph neural network-based greedy exploration policy with a safety filter that replaces unsafe actions with the closest feasible alternative. Second, we design a graph-based map representation together with an attention-based graph neural network for the implementation of the exploration policy. Third, we propose a reward function that promotes map expansion through potential field variations driven by frontier information gain and agent--frontier distance. The proposed framework is extensively validated through benchmark comparisons.

\begin{figure*}[h]
    \centering
    \includegraphics[width=\linewidth]{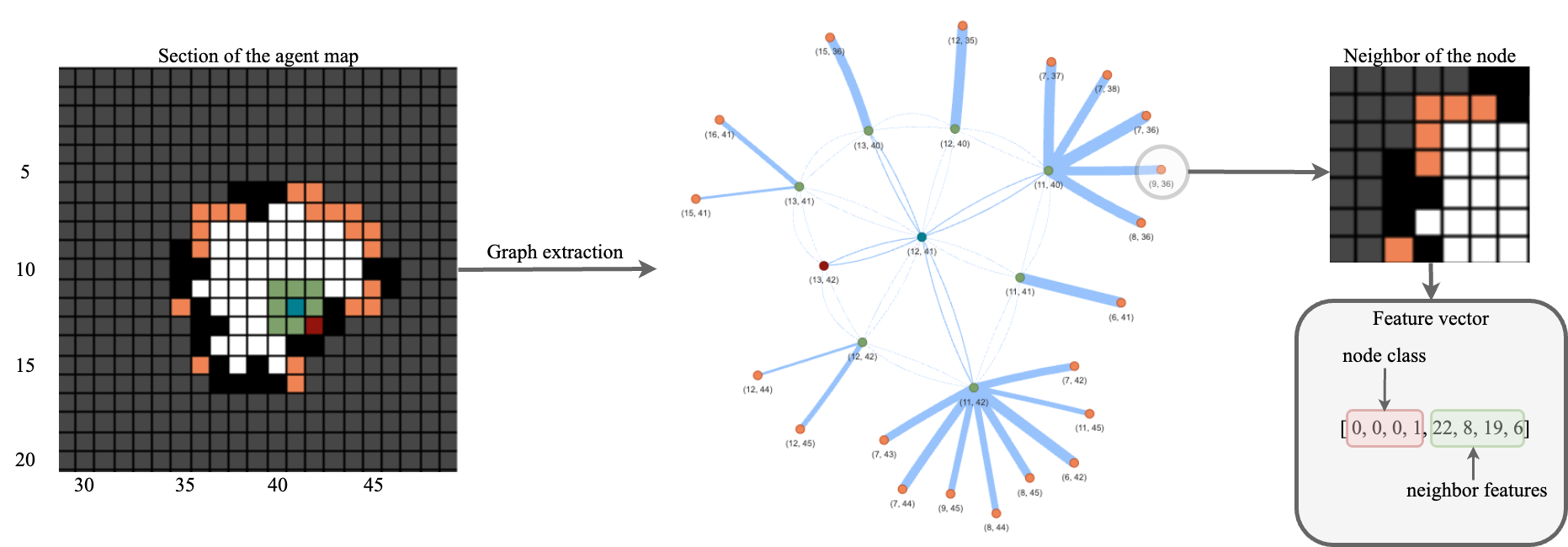} 
    \caption{Illustration of the exploration graph extraction, used as observation $o_t$, representing the agent's exploration map. The image depicts a section of the known arena with occupied (black), unknown (gray), and free (white) cells. The agent's position, feasible and unfeasible next-step navigation goals, and frontiers are marked in blue, green, red, and orange, respectively. The graph structure captures node relationships, while the right side highlights a frontier’s local neighborhood and its extracted feature vector. Moreover, the graph's edge thickness is proportional to the distance between connected nodes. }
    \label{fig:exploration_graph}
\end{figure*}

\section{Methodology}
\label{sec:methodology}

\subsection{Reinforcement learning formulation}
\label{sbs:agent_policy}

\begin{table}[t]
\centering
\caption{Displacements $\Delta$ associated to each action in $\mathcal{A}$}
\label{tab:action_space}
\begin{tabular}{cccccc}
\toprule
Action & $\Delta$ & Action & $\Delta$ & Action & $\Delta$ \\
\midrule
$a^{(0)}$ & $(0,0)$   & $a^{(3)}$ & $(-1,1)$ & $a^{(6)}$ & $(1,-1)$ \\
$a^{(1)}$ & $(-1,-1)$ & $a^{(4)}$ & $(0,-1)$ & $a^{(7)}$ & $(1,0)$ \\
$a^{(2)}$ & $(-1,0)$  & $a^{(5)}$ & $(0,1)$  & $a^{(8)}$ & $(1,1)$ \\
\bottomrule
\end{tabular}
\end{table}

\begin{algorithm}[t]
\caption{Implementation of the safety filter}
\label{alg:safety_filter_high_level}
\begin{algorithmic}[1]
\Require Current position $\mathbf{p}^t$, policy-selected action $u_t^\pi$, action set $\mathcal{A}$
\Ensure Safe action $u_t^\sigma$
\State \textbf{Definition:} $\operatorname{idx}(a^{(i)}) = i$, for $i \in \{0,\ldots,8\}$

\If{$u_t^\pi$ is feasible}
    \State $u_t^\sigma \gets u_t^\pi$
\Else
    \State $\mathcal{A}_t^f \gets \{a' \in \mathcal{A}\setminus\{a^{(0)}\} \mid a' \text{ is feasible}\}$
    \If{$\mathcal{A}_t^f = \emptyset$}
        \State $u_t^\sigma \gets a^{(0)}$
    \Else
        \State $u_t^\sigma \gets \arg\min\limits_{a' \in \mathcal{A}_t^f} \left|\operatorname{idx}(a') - \operatorname{idx}(u_t^\pi)\right|$
    \EndIf
\EndIf
\end{algorithmic}
\end{algorithm}

An overview of the proposed framework is presented in Fig. \ref{fig:framework_overview}. In particular, the exploration arena is represented by an occupancy grid map $\mathcal{M}_{h \times h}$, which encodes the spatial distribution of $n_t$ circular-shaped non-traversable regions with radius $r_t$ corresponding to tree trunks in a planar section parallel to the ground. At each time step $t$ the exploring agent occupies the grid $\mathbf{p}^{t} = [\,p_x^{t},\; p_y^{t}\,]^\top$ and has its map of the environment $\mathcal{M}^a_t$ which includes only the explored regions, with all other cells marked as unknown. When the agent reaches a goal, it updates $\mathcal{M}^a_t$ with data from a simulated 360-degree LiDAR sensor with range $l$. A graph-based representation of the explored environment $o_t$ is then computed for the agent's policy, incorporating the agent's position, navigation goals, and the border between traversable and unknown regions, as detailed in \ref{sbs:graph_based_observation}. As in Fig. \ref{fig:framework_overview}, the proposed agent's policy combines an exploration greedy policy $\pi_\theta$ that select a movement $u_t^\pi \in \mathcal{A} \sim \pi_\theta(\cdot \mid o_t)$ towards one of the eight neighboring cells and a safety filter $\sigma(u^\pi_t)$ hard-constrained to prevent unfeasible actions. In particular, the action space $\mathcal{A} = \left\{ a^{(0)}, \ldots, a^{(8)} \right\}$ is defined as per Table \ref{tab:action_space}. The safety filter, whose implementation is shown in Algorithm \ref{alg:safety_filter_high_level}, evaluates the policy-selected action $u_t^\pi$ based on the agent's current position $\mathbf{p}^{t}$ and observation $o_t$. An action is considered feasible if the displacement induced by the action leads to a position that remains within the environment bounds and does not collide with an obstacle. If $u_t^\pi$ is feasible, the filter accepts it unchanged. Otherwise, it constructs the feasible action set $\mathcal{A}_t^f \subseteq \mathcal{A}$, containing all admissible actions except the null action. If this set is non-empty, the filter selects the feasible action that is closest to $u_t^\pi$ according to the action ordering; otherwise, it returns the null action $a^{(0)}$. The GNN-based policy $\pi_\theta$ is trained via reinforcement learning to maximize the expansion of $\mathcal{M}^a_t$ while minimizing unfeasible actions $u^\pi_t$ that require intervention from the safety filter. The action $u^\sigma_t$ is then used to move the agent to a new location, and this process repeats until either the exploration ratio \( \rho_t \geq \rho^* \) or the number of iterations \(n_s \geq n_s^* \), where $\rho^* $ and $n_s^* $ are the exploration threshold and the maximum number of the agent's interactions with the environment.

Moreover, the exploration greedy policy is trained using the Policy Proximal Optimization (PPO)\cite{schulman2017proximalpolicyoptimizationalgorithms} algorithm and a critic network $V_{\phi}$ is included as described in \ref{sbs:gnn_architecture}. At each step, the reward \( r(s_t, u^\pi_t, u^\sigma_t, s_{t+1}) \), where \( s_t \) and \( s_{t+1} \) represent the agent's position and exploration discoveries before and after action \( u^\sigma_t \), is generated by the environment.

\subsection{Graph-based agent's observation}
\label{sbs:graph_based_observation}

The agent's observation \( o_t \) at time \( t \) is an exploration graph \( \mathcal{G} = (\mathcal{V}, \mathcal{E}) \), where nodes \( \mathcal{V} \) represent the agent's position \( v^a_t \), neighboring cells $\mathcal{V}_n$, and frontiers $\mathcal{V}_f$, while edges \( \mathcal{E} \) encode spatial relations with both undirected and bidirected connections, as shown in Fig. \ref{fig:exploration_graph}. In particular, $\mathcal{E}$ includes edges bi-connecting the agent node $v_a$ to each possible next-step waypoint $v_n \in \mathcal{V}_n$. Moreover, bidirectional connections between the nodes in $\mathcal{V}_n$ are also added according to Eq. \eqref{eq:closeneig}.

\begin{equation}
\label{eq:closeneig}
    \mathcal{E}_n = \{ \langle v_n, v_m \rangle \mid v_n, v_m \in \mathcal{V}_n, \, \| v_n - v_m \|_\infty \leq 1 \}
\end{equation}

where $\| \cdot \|_\infty$ is the Chebyshev norm. Furthermore, each frontier node \( v_f \in \mathcal{V}_f \) is connected to its nearest navigation node \( v_n^* \), linking unexplored regions to waypoints according to $\mathcal{E}_f = \{ (v_f, v_n^*) \mid v_f \in \mathcal{V}_f \}$ where $v_n^*$ is given by Eq. \ref{eq:closest_nav}.

\begin{equation}
\label{eq:closest_nav}
v_n^* = \arg \min_{v_n \in \mathcal{V}_n} \| v_f - v_n \|_2,
\end{equation}  
Each edge is weighted by the Euclidean distance between its nodes, and each node \( v_i \) has a feature vector \( \mathbf{x}_{v_i} \in \mathbb{R}^8 \) encoding the node class and local environment features extracted from the agent's map $\mathcal{M}^a_t$. In particular, the first four dimensions of the node feature vector use one-hot encoding, a method for converting categorical variables into a binary format, to define if it is the agent node, a traversable or non-traversable next-step navigation node or a frontier node.  Furthermore, each node \( v_i \) encodes local occupancy statistics from a \( k \times k \) neighborhood, summarizing the number of free, unknown, occupied, and frontier cells around the node $v_i$ according to the agent's map $\mathcal{M}^a_t$ as illustrated in Fig. \ref{fig:exploration_graph}. 

\subsection{Architecture of the greedy exploration policy and critic}\label{sbs:gnn_architecture}

\begin{table}[b]
\centering
\caption{Hyperparameters of the GATv2Conv layers}
\label{tab:gnn_architecture}
\begin{tabular}{lcccc}
\toprule
\textbf{Layer} & \textbf{Input dim.} & \textbf{Output dim.} & \textbf{Heads} & \textbf{Activation} \\
\midrule
GATv2Conv 1 & 8  & 16 & 4 & ReLU \\
GATv2Conv 2 & 64 & 1  & 1 & None \\
\bottomrule
\end{tabular}
\end{table}

Both the exploration greedy policy $\pi_\theta$ and the critic network $V_\phi$ are implemented as multi-head attention-based graph neural networks leveraging two sequential GATv2Conv \cite{brody2022attentivegraphattentionnetworks} layers that use attention mechanisms to compute node representations by considering both node features and edges. In particular, the first layer has hyperparameters shown in Table \ref{tab:gnn_architecture} and 4 parallel attention heads to transform node features into a concatenated higher-dimensional hidden representation according to Eq. \eqref{eq:node_feat_ev}.  

\begin{equation}
\label{eq:node_feat_ev}
    \mathbf{x}_{v_i}^{(1)} = \oplus_{h=1}^{4}\left( \sum_{v_i \in \mathcal{N}(v_i) \cup \{ v_i \}}
    \alpha_{v_i,v_j}^{h} \mathbf{\Theta}_{t}^{h} \mathbf{x}_{v_j}\right)
\end{equation}

where \( \mathbf{\Theta}_t^{h}\) is the learned transformation matrix for target nodes from head $h$, $\mathcal{N}(v_i)$ denotes the neighbors of node \( v_i \), $\oplus$ stands for the concatenation operator and the adaptive attention coefficients $\alpha_{v_i,v_j}^{h}$ are computed as per Eq. \eqref{eq:adaptive_attention_coefficients}. 

\begin{equation}
\label{eq:adaptive_attention_coefficients}
    \alpha_{v_i,v_j}^{h} =
    \frac{
    e^{\boldsymbol{\epsilon}_{v_j}^h}}
    {\sum_{v_k \in \mathcal{N}(v_i) \cup \{ v_i \}}
    e^{\boldsymbol{\epsilon}_{v_k}^h}}
\end{equation}

where $\boldsymbol{\epsilon}^{h}_{v_w} =\mathbf{a}^h\top\mathrm{LeakyReLU}\left(
    \mathbf{\Theta}_{s}^{h}\mathbf{x}_{v_i} + \mathbf{\Theta}_{t}^{h}\mathbf{x}_{v_w}
    \right)$,  \( \mathbf{\Theta}_s^{h}\) is the transformation matrix for source nodes, and \( \mathbf{a}^h \) is the learnable attention vector.
A second GATv2 layer refines these embeddings using a single attention head producing the embedding $\mathbf{x}_{v_i}^{(2)}$ as per Eq. \eqref{eq:second_emb}.  

\begin{equation}
\label{eq:second_emb}
    \mathbf{x}_{v_i}^{(2)} = \sum_{v_j \in \mathcal{N}(v_i) \cup \{ v_i \}} \alpha_{v_i,v_j} \mathbf{W} \mathbf{x}_{v_j}^{(1)}
\end{equation}

where \(\mathbf{W}\) is the learned projection matrix.  

The exploration policy and critic network both process the node features \( v_{out} = \{ v_a \} \cup \mathcal{V}_n \) representing the learned ranking of candidate waypoints. The policy outputs the features of these nodes, while the critic aggregates the features and applies a fully connected layer.
\subsection{Reward shaping with frontier-based potential field} 
\label{sbs:reward}

The reward function used to train the exploration greedy policy is formulated to encourage efficient exploration and penalize the intervention of the safety filter requested to avoid collisions. Therefore, the reward \( r_t(s_t, u^\pi_t, u^\sigma_t, s_{t+1}) \) is defined according to Eq. \eqref{eq:reward}:

\begin{equation}
\label{eq:reward}
r_t = 
\begin{cases} 
r_{exp}, & \text{if } \rho_t \geq \rho^* \text{ and } u^\pi_t = u^\sigma_t \\[8pt]
n^e_t + \Phi(s_{t+1}) - \Phi(s_t), & \text{if } \rho_t < \rho^* \text{ and } u^\pi_t = u^\sigma_t \\[8pt]
r_\sigma, & \text{if } u^\pi_t \neq u^\sigma_t
\end{cases}
\end{equation}

where $n^e_t$ is the number of explored cells added to the agent's map at time step $t$ due to the execution of $u^\sigma_t$, and $\Phi(\cdot)$ is the function computing the frontier-based potential field. This map evaluates the current state of exploration $s_t$ by considering both the spatial distances of the frontiers from the agent position $\mathbf{p}^{t}$ and the expected information gain of these frontiers. In particular, given a set of frontiers \(\mathcal{F}_t = \{f_1, f_2, ..., f_n\}\), where each \(f_i\) represents the coordinates of a frontier, the distances between the current position of the agent \(\mathbf{p}^{t}\) and each frontier are computed as per Eq. \eqref{eq:potential_distances}.

\begin{equation}
\label{eq:potential_distances}
    d_i = \|f_i - \mathbf{p}^{t}\|_2 
\end{equation}
and $\bar{d}_i$ corresponds to the value of $d_i$ scaled in the range $[0,1]$.
The expected information gain for each frontier \( f_i \) is then computed based on the number of unknown grid cells in a local region $\mathcal{N}$ around each frontier of size $k \times k$ as per Eq. \eqref{eq:expected_inf_gain}.

\begin{equation}
\label{eq:expected_inf_gain}
    \Gamma_i =  \sum_{x \in \mathcal{N}(f_i)} \gamma(x) 
\end{equation}

where \(\gamma(\cdot)\) is the function that counts unknown cells in the neighbor $\mathcal{N}$, and $\bar{\Gamma}_i$ corresponds to the value of ${\Gamma}_i$ scaled in the range $[0,1]$. 
Then the potential field $\Phi_i$ associated with each frontier is computed according to Eq. \eqref{eq:potential_frontier} as a linear combination of the distances and information gains and the trade-off between these two factors is controlled by a parameter \(\beta\).

\begin{equation}
   \Phi_i = -\bar{d}_i + \beta \cdot \bar{\Gamma}_i \label{eq:potential_frontier} 
\end{equation}

The potentials are then normalized again to ensure they lie within the range \([0, 1]\). Finally, the frontier potential in state $s_t$ is determined as the maximum potential across all frontiers.
\section{Simulation}

\begin{figure*}[t]
    \centering
    \includegraphics[width=\textwidth]{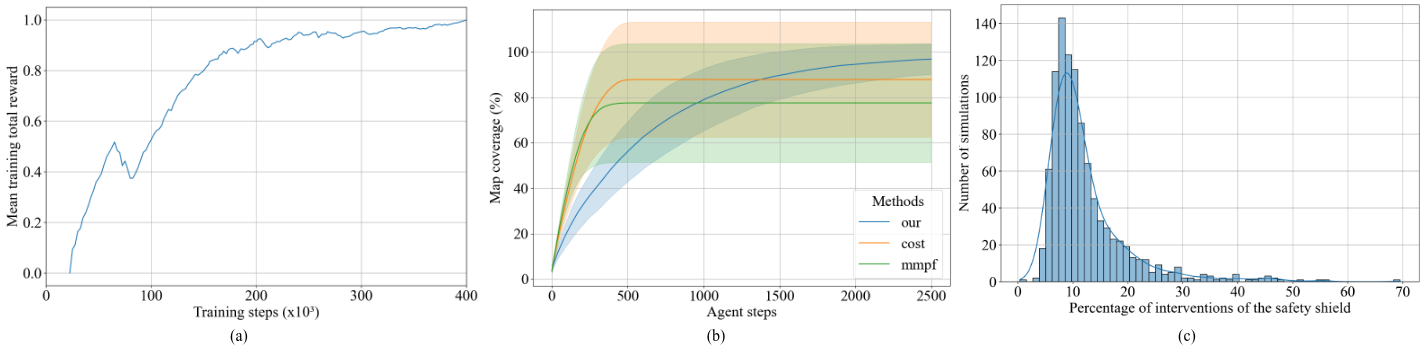}
    \caption{From left to right: (a) Mean total reward curve during training, smoothed and scaled to $[0,1]$. (b) Map coverage percentage relative to traversable regions over agent steps in 1000 test environments considering the proposed method and two benchmarks. (c) Distribution of the percentages of safety filter interventions over agent actions across 1000 test simulations.}
    \label{fig:nom}
\end{figure*}

\begin{figure*}[b]
    \centering
    \includegraphics[width=\textwidth]{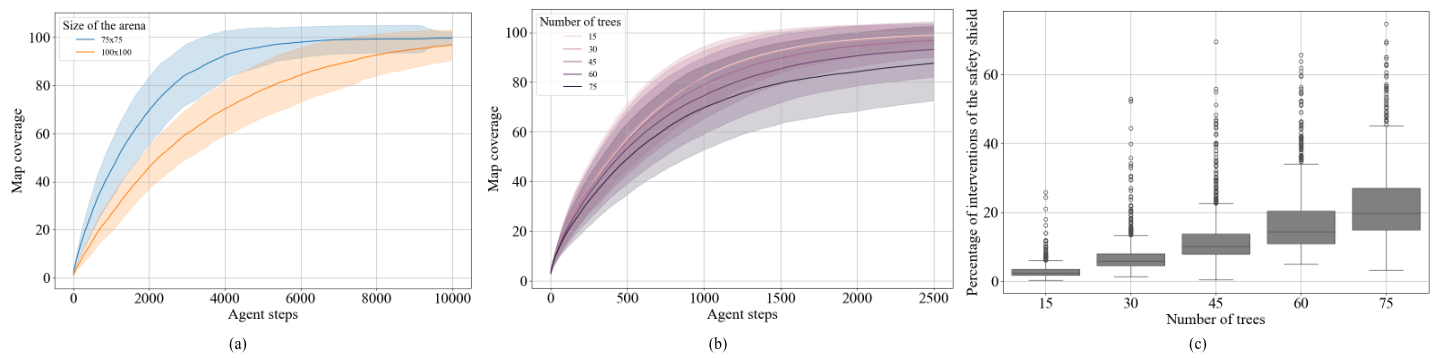}
    \caption{From left to right: (a)–(b) Map coverage percentage over agent steps in environments with varying sizes of the exploration arena and non-traversable regions. (c) Distribution of safety filter intervention ratio across agent actions for different tree densities.}
    \label{fig:rob}
\end{figure*} 

\subsection{Training setup}\label{sbs:training}

\begin{table}[t]
\centering
\caption{Environment and training hyperparameters}
\label{tab_parameters}
\begin{tabular}{l c l c}
\toprule
\multicolumn{2}{c}{\textbf{Environment}} & \multicolumn{2}{c}{\textbf{Training}} \\
\cmidrule(r){1-2} \cmidrule(l){3-4}
\textbf{Parameter} & \textbf{Value} & \textbf{Parameter} & \textbf{Value} \\
\midrule
$h$         & $50$                & Rollouts                    & $1024$         \\
$n_t$       & $45$                & Mini-batches                & $64$           \\
$r_t$       & $[1,3]$             & Learning rate               & $3 \times 10^{-4}$ \\
$l$         & $5.0$               & Learning epochs             & $8$            \\
$\rho^{*}$  & $0.98$              & Discount factor $\gamma$    & $0.99$         \\
$n_s^{*}$   & $2500$              & Training timesteps          & $4 \times 10^{5}$ \\
$k$         & $7$                 & Clipping parameter $\epsilon$ & $0.2$        \\
\bottomrule
\end{tabular}
\end{table}

The training was conducted on HPC2N's Kebnekaise cluster running Ubuntu 20.04.6 LTS, utilizing 5 Skylake CPU cores and an NVIDIA H100 GPU, while policy evaluations were performed on the same system using 10 Skylake CPU cores. The environment was implemented using Gymnasium \cite{towers2024gymnasium}, with the key design parameters summarized in Table \ref{tab_parameters}. The exploration policy and critic were modeled using a graph-based neural network implemented in PyTorch Geometric \cite{Fey/Lenssen/2019}. Training was performed in environments with varying non-traversable region distributions and initial agent placements to enhance policy generalization. The PPO algorithm was implemented using skrl \cite{serrano2023skrl}, with the key hyperparameters detailed in Table \ref{tab_parameters}. Moreover, the reward function is designed considering $r_{exp} = 100$, $r_\sigma = -5$ and $\beta = 0.5$.
Fig. \ref{fig:nom}-(a) shows the mean total reward curve collected during the training after having performed a smoothing and a scaling of the obtained rewards in the range $[0, 1]$. It can be seen that the rewards generally increase except for a small decline after $6 \times 10^4$ steps from the beginning of the training. Nevertheless, the graph demonstrates that the reward curve converges.

\subsection{Simulation results}\label{sbs:results}
The policy trained with the parameters in Table \ref{tab_parameters} is evaluated in 1000 test environments with randomly placed non-traversable regions and agent starting positions. The framework, consisting of an RL-based greedy exploration policy and a safety filter, is executed for 2500 time steps to assess its robustness across different environment configurations. Figure \ref{fig:nom}-(b) shows the average map coverage over agent steps during exploration for our proposed method and the benchmark approaches implemented by \cite{9812344}. Specifically, the cost-based method selects the nearest frontier to minimize travel distance, while the MMPF approach employs potential fields to guide the robot smoothly toward informative frontiers and reduce redundant motion. The original implementations have been used, except that the LiDAR range has been adjusted to match the value reported in Table \ref{tab_parameters}. Considering the proposed approach, the results show a steady increase in coverage, reaching 80\% after 1000 steps and approximately 95\% after 2000 steps. This demonstrates the framework’s capability to achieve high exploration efficiency despite variations in environment configurations. While the methods implemented in \cite{9812344} achieve only 88\% and 78\% coverage, our approach continues to explore effectively, reaching nearly 97\% average map coverage. Additionally, Fig. \ref{fig:nom}-(c) presents the percentage of the number of steps in which the safety filter intervenes to avoid a collision over the total number of actions per simulation, showing that in 80\% of simulations, interventions occur in less than 15\% of steps, with an average intervention rate of around 10\%. This demonstrates that the policy effectively explores with minimal reliance on the safety filter. However, its utility is evident during execution, as it is occasionally activated when necessary. To further analyze robustness, the proposed approach is tested in environments with varying numbers of trees and different map sizes. In the first case, 1000 simulations are conducted, each running for 2500 steps, while in the second case, 100 simulations are executed for 10,000 steps. In particular, Fig. \ref{fig:rob}-(a) demonstrates the policy’s adaptability to different environment sizes. The agent successfully explores more than 80\% of the traversable area after approximately 2500 steps in maps 125\% larger than the nominal size and after 5000 steps in maps 300\% larger. Moreover, Fig. \ref{fig:rob}-(b) shows that while an increase in non-traversable regions slightly reduces map coverage considering the same number of agent steps, the policy still achieves over 80\% coverage even with a 67\% increase in the number of trees. The increased obstacle complexity is also reflected in Fig. \ref{fig:rob}-(c), where the safety filter activates more frequently as the number of trees grows. This highlights the necessity of the safety module in ensuring safe exploration. However, even in the most complex configurations, the policy maintains the intervention rate around 20\% of the total exploration steps. These results confirm the framework’s scalability and robustness in diverse exploration scenarios.

\section{CONCLUSIONS}
This paper presents a framework for safe and efficient exploration in cluttered environments. The method uses an attention-based graph neural network trained with reinforcement learning, together with a graph-based representation of relevant locations and a frontier-based potential field reward to favor informative nearby frontiers while reducing safety shield interventions. Simulations against benchmark methods show robust map coverage across diverse environments, with safety filter interventions below 
10\%. Robustness is further confirmed under varying tree densities and environment sizes. Future work will focus on real-world deployment as a high-level exploration planner for an unmanned aerial vehicle in forest environments.



\bibliographystyle{IEEEtran}
\bibliography{main}

\end{document}